\documentclass[conference]{IEEEtran}
\IEEEoverridecommandlockouts

\usepackage[numbers]{natbib}
\usepackage{amsmath,amssymb}
\usepackage{epsfig}
\usepackage{graphicx}
\usepackage{tabularx}

\usepackage{mathtools}

\usepackage[T1]{fontenc}
\usepackage[english]{babel}
\usepackage[utf8]{inputenc}

\def\BibTeX{{\rm B\kern-.05em{\sc i\kern-.025em b}\kern-.08em
    T\kern-.1667em\lower.7ex\hbox{E}\kern-.125emX}}

\def\real{\mathcal R}
\def\gd{\text{d}}

\DeclarePairedDelimiter\floor{\lfloor}{\rfloor}

\usepackage[pagebackref=false,breaklinks=true,letterpaper=true,colorlinks,bookmarks=false]{hyperref}

\begin{document}

\title{DCT-Conv: Coding filters in convolutional networks \\ with Discrete Cosine Transform}

\author{\IEEEauthorblockN{Karol Chęciński, Pawe\l{} Wawrzy\'nski}
 \IEEEauthorblockA{\textit{Institute of Computer Science}\\
  \textit{Warsaw University of Technology}\\
  Nowowiejska 15/19, 00-665 Warsaw, Poland \\
  karol.checinski.stud@pw.edu.pl, pawel.wawrzynski@pw.edu.pl
 }
}

\maketitle

\begin{abstract}
Convolutional neural networks are based on a~huge number of trained weights. Consequently, they are often data-greedy, sensitive to overtraining, and learn slowly. We follow the line of research in which filters of convolutional neural layers are determined on the basis of a~smaller number of trained parameters. In this paper, the trained parameters define a~frequency spectrum which is transformed into convolutional filters with Inverse Discrete Cosine Transform (IDCT, the same is applied in decompression from JPEG). We analyze how switching off selected components of the spectra, thereby reducing the number of trained weights of the network, affects its performance. Our experiments show that coding the filters with trained DCT parameters leads to improvement over traditional convolution. Also, the performance of the networks modified this way decreases very slowly with the increasing extent of switching off these parameters. In some experiments, a~good performance is observed when even $99.9\%$ of these parameters are switched off. 
\end{abstract}



\section{Introduction}

Convolutional neural networks are now the most efficient tool for image analysis tasks such as face recognition \cite{6909616}, medical diagnostic \cite{2017Esteva,8050011} or objects detection \cite{yolo}. Contemporary models such as VGG \cite{1409.1556}, ResNet \cite{1512.03385}, GPipe \cite{1811.06965} have up to hundreds of millions of trained parameters. Training them requires huge volumes of data or it is prone to overtraining and poor behavior on unseen images. 

A~way to avoid the aforementioned unfortunate alternative is as follows: For each filter (or each filter's part) in the network there exists a~vector of trained parameters; its dimension is smaller than the size of the filter (or its part); a~filter (or its part) is produced from the trained parameters with a~certain fixed transformation. Fewer trained parameters represent a~similar diversity of filters. The above fixed transformation is a~key element that needs to be designed for this approach to be successful. 

In this study we analyze the aforementioned transformation based on (Inverse) Discrete Cosine Transform, (I)DCT. DCT can be applied to a~matrix $m \times n$ of real numbers to produce a~matrix of the same size of real numbers --- DCT coefficients. The matrix of DCT coefficients can be transformed back, with IDCT to the original matrix. For the original matrix which is a~part of a~real image, often some DCT coefficients are close to zero. Therefore, JPEG compression is based on the DCT of parts of the image with skipping DCT coefficients that are close to zero. 

The contribution of this paper can be summarized in the following points: 
\begin{itemize} 
\item We propose to train DCT coefficients of filters in convolutional neural networks rather than the filters themselves. 
\item We demonstrate on four large learning problems that the above modification causes a~significant improvement in the performance of the networks on the test data. 
\item We analyze how the performance of the above networks is influenced by switching off the DCT coefficients of the filters, i.e. setting them equal to zero and excluding them from training. We demonstrate that the performance decreases very slowly with increasing the extent of the switching off.
\end{itemize} 
The remainder of the paper is organized as follows. The next section presents related work. For this paper to be self-contained, Sec.~\ref{sec:dct} presents DCT, IDCT, and some of their properties. Sec.~\ref{sec:method} presents the DCT-Conv layer, a~layer in which filters are defined by trained DCT coefficients. Sec.~\ref{sec:experiments} presents an empirical study. The last section concludes the paper. 


\section{Related work}
\label{sec:related:work} 

This paper is related to two issues elaborated in the literature: The first one is how to reduce the number of trained parameters in neural networks. The second one is the application of the DCT in the processing of data by neural networks. 


\subsection{Reduction of trained parameters} 


There are two main motives to reduce the number of trained parameters in a~neural network. The first one is a~better generalization, thus a~lower demand for data. The second one is a~lower demand for memory to store those parameters which enables broader applicability, e.g. in mobile devices and embedded systems. 


In \cite{1510.00149} a~method of pruning parameters of small absolute value was introduced that enabled reducing the number of trained parameters from AlexNet by 85\% without accuracy deterioration. In \cite{Denil:2013:PPD:2999792.2999852}, 75\% of the parameters were reduced in a~convolutional network due to the decomposition of the filter matrix. In \cite{pmlr-v37-chenc15} hashing functions were applied to divide parameters into bins in which they are trained together. \cite{tan2019efficientnet} shows that a~significant number of parameters can be reduced by the appropriate shaping of the input and a~network structure. 


An~approach to parameters' reduction in convolutional networks is the application of the same filters in different scales \cite{1411.6369} and rotated by different angles \cite{Marcos_2017_ICCV}. This way a~network does not need to learn multiple filters to represent the same pattern at different scales and angles. Consequently, different filters may cover better all the patterns present in the data that the network needs to be able to identify. 


\subsection{Application of Discrete Cosine Transform in neural networks} 


DCT has been applied to process input images for convolutional neural networks. In \cite{Ulicny2017OnUC} this approach was combined with the transformation of images to YCbCr color space and applied to image classification and verified on MNIST and CIFAR-10 dataset. Also, images transformed to the frequency domain are analyzed in specific applications. E.g. in \cite{LeukoNet}, features of images are produced by convolutional networks processing both original images and their frequency representations. That architecture was applied to the classification of cells suspected of acute lymphoblastic leukemia (ALL). DCT was applied to compress filters in convolutional networks in \cite{cnnpack}.



\section{Discrete Cosine Transform}
\label{sec:dct} 

Let $x \in \real^{m\times n}$ be a~matrix with entries $x_{i,j}, {i=0,...,m-1}, j=0,...,n-1$. DCT-II (Discrete Cosine Transform type II)\footnote{There are several types of DCT; type II is the most popular one in multimedia.} of $x$ produces a~matrix, $X\in\real^{m\times n}$, whose entries, DCT coefficients, $X_{k,l}, k=0,...,m-1, l=0,...,n-1$ are given by the formula: 
\begin{equation} 
    \begin{split} 
    X_{k,l} = \sum_{i=0}^{n-1} \sum_{j=0}^{m-1} x_{i,j} 
    & \cos\left[\frac\pi{n}\left(j + \frac12\right)l\right] \times \\
    & \times \cos\left[\frac\pi{m}\left(i + \frac12\right) k\right].
    \end{split} \label{def:DCT} 
\end{equation} 
Given DCT coefficients $X$, the original image $x$ can be reconstructed with IDCT-II (Inverse Discrete Cosine Transform type II) as follows 
\begin{equation}
    \begin{split} 
    x_{i,j} = \sum_{k=0}^{m-1}\sum_{l=0}^{n-1} X_{k,l} 
    & \cos\left[\frac\pi{n}\left(j+\frac12\right)l\right] \times \\
    & \times \cos\left[\frac\pi{m}\left(i+\frac12\right)k\right]. 
    \end{split} \label{def:IDCT}
\end{equation}
The formulae for DCT-II and IDCT-II are thus very similar. They apply to matrices. Formulae for DCT-II and IDCT-II for vectors are easily obtained by setting above $n=1$. Formulae for DCT-II and IDCT-II for tensors of higher order are analogical to those above. 

In words, the DCT coefficient $X_{k,l}$ says how strong in $x$ the component is whose frequency over the first coordinate is proportional to $k$, and whose frequency over the second coordinate is proportional to $l$. In particular, $X_{0,0}$ is the sum of all elements in $x$. 

DCT is applied in JPEG compression: An image is divided into squares and their DCT is computed. Only DCT coefficients with sufficiently large absolute values are put into the compressed file. Usually, that means that high frequency components that correspond to noise and small details in the image, insignificant for the viewer anyway, are skipped. 



\section{Method}
\label{sec:method} 

The general idea analyzed in this paper is as follows. We take a~convolutional layer of a~neural network. Whenever the filters in the layer are in use, they result from IDCT on trained weights, instead of being trained parameters themselves. Specifically, tensors with trained weights are transformed into filters of the same shape through IDCT. However, some of the weights are set to zero. Hence, there are fewer trained parameters than in the original convolutional layer, but the filters within the layer are still diverse enough to respond to different patterns in data. 


\subsection{DCT-Conv layer}

Trained parameters (weights) of a~DCT-Conv layer create a 4th-order tensor in $\real^{N\times C\times H\times W}$, where $N$ is a~number of filters, $C$ is a~number of channels, and $H,W$ are the height and width of a~filter, respectively. A~filter is a~sequence of $C$ slices -- $H\times W$ matrices. IDCT is performed on the weights tensor to create a tensor of filters of the same shape, such that IDCT is performed independently for each slice. Once created, the~tensor of filters is used as usual in a~convolutional layer. 


In some architectures, like ResNet, there are filters with $W=H=1$. Effectively, a~layer of such filters is defined by a~matrix in $\real^{N\times C}$ (tensor of order 2). While this matrix could also be computed on the basis of its trained DCT coefficients, this is problematic because of two reasons. Firstly, while one may expect a~certain regularity in relation between values in filters and their spacial coordinates, existence of a~relation between values in filters and indices of channel and especially filter is less obvious. Secondly, usually $N,C \gg W,H$ which makes IDCT of the aforementioned matrix expensive. Because of these problems, in the experiments discussed below a~matrix of $1\times1$ filters is produced with IDCT separately for its subrows of length 16. 

In order to run any gradient-based learning algorithm with a~network that contains a~DCT-Conv layer, derivatives of the cost function need to be computed with respect to the trained weights. Let us denote the cost function by $J$, and suppose that its derivatives with respect to filter components, denoted by $x_{i,j}$ \eqref{def:IDCT}, namely $\gd J/\gd x_{i,j}$, are known. Then, the derivatives of the cost function with respect to the trained weights, denoted by $X_{k,l}$, are determined through error backpropagation i.e., by analyzing how $X_{k,l}$ influences different $x_{i,j}$ in \eqref{def:IDCT}, namely
\begin{equation}
    \begin{split} 
    \frac{\gd J}{\gd X_{k,l}} = \sum_{i=0}^{m-1}\sum_{j=0}^{n-1} \frac{\gd J}{\gd x_{i,j}} 
    & \cos\left[\frac\pi{n}\left(j+\frac12\right)l\right] \times \\
    & \times \cos\left[\frac\pi{m}\left(i+\frac12\right)k\right]. 
    \end{split} \label{def:d:IDCT}
\end{equation}

Some of the weights are switched off i.e., set to zero and excluded from training. They are randomly, independently between one another selected before the training. The switch-off probability is defined by a~coefficient, $p\in[0,1]$. 

\begin{figure}[t]
\begin{center}
\includegraphics[width=0.8\linewidth]{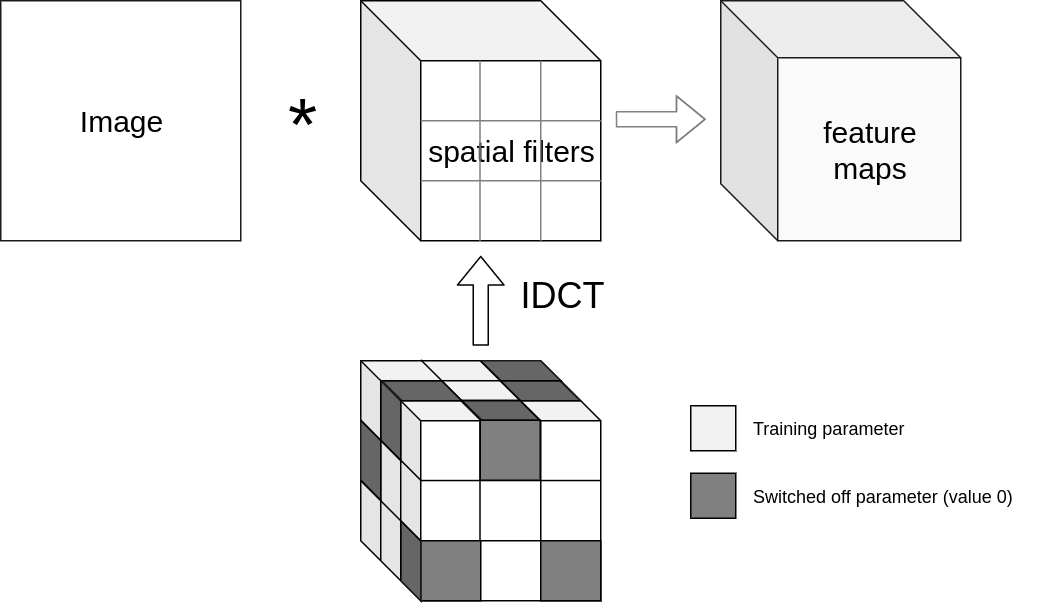}
\end{center}
    \caption{DCT-Conv layer. Gray weights are switched off (set to zero). }
    \label{fig:dctlayer}
\end{figure}

While in the future we plan to research more efficient schemes of switching the weights off, now we only analyze the~above simple scheme. The DCT-conv layer is presented in Fig.~\ref{fig:dctlayer}. 


\section{Experiments}
\label{sec:experiments} 

In the empirical study below we compare the performance of original networks with convolutional layers against similar architectures with DCT-Conv layers with the same number of filters of the same size. Also, we analyze how the switch-off probability impacts the performance of the networks with DCT-Conv layers. We analyze 4 benchmark learning problems, all based on the CIFAR-100 dataset of images.

\subsection{Dataset: CIFAR-100} 
\label{sec:cifar100} 

The CIFAR-100 dataset \cite{cifar}  contains 60,000 images in RGB with a~size of 32x32px. Each image is of one of 100 equivocal disjoint classes. The dataset is divided into a~training set (50,000 images) and a~test set (10,000 images). 

In our experiments the images have been normalized: For each channel, $c$, an average, $av(I_{tr}[\cdot,\cdot,c])$, and standard deviation, $\sigma(I_{tr}[\cdot,\cdot,c])$ were computed for the images in the training set. The normalized images feeding the neural networks are computed according to
\begin{equation} 
    \hat{I}[i,j,c] = (I[i,j,c] - av(I_{tr}[\cdot,\cdot,c]) / \sigma(I_{tr}[\cdot,\cdot,c]) 
\end{equation}
where $(i,j)$ are the coordinates. 


\subsection{Problem 1. ResNet50 classifier} 
\label{sec:resnet-cifar100} 

\paragraph{Architecture} 
We adopt the ResNet50 architecture with bottleneck blocks \cite{1512.03385}. ResNets are generally deep modular convolutional networks parameterized by the~number of layers. Characteristic modules in ResNets are residual blocks that combine convolution with parallel passing of the block input to its output. These by-passes prevent exploding and vanishing gradients, and enable very deep architectures. 


The orthogonal initialization \cite{orthogonal} has been applied to all the filters within the network. The output dense layer has been initialized by means of Glorot's method \cite{pmlr-v9-glorot10a}. 


The network architecture is depicted in detail in~Tab.~\ref{tab:ResNet50}. The network has 23,676,990 trained parameters, including 11,317,248 in 3x3 filters, and 12,130,384 in 1x1 filters. 


\paragraph{Training} 

Classic Momentum \cite{Qian:1999:MTG:307343.307376} was applied with a~momentum decay factor of 0.9. The step-size was evolving in the training time according to the following scheme: 0.001 in epoch 1, 0.1 till epoch 60, 0.02 till epoch 120, 0.004 till epoch 160, and 0.0008 till the final epoch 200. The minibatches of size 128 were used. 

\begin{table}
    \caption{ResNet50 architecture. ConvBlock2D and IdentBlock2D are presented in Fig.~\ref{fig:resblocks}. The numbers in the Size column next to blocks denote the numbers of filters within convolutional layers in a~block. Column 3 and 4: number of parameters in $3\times3$ and $1\times1$ filters, respectively.  } 
    \label{tab:ResNet50}
    \centering
    \begin{tabular}{|c|c|c|c|} 
    \hline
    \multicolumn{4}{|c|}{ResNet50} \\ \hline \hline
    Layer/Block & Size & \multicolumn{2}{|c|}{Parameters} \\ \hline
    Conv2D & 64x3x3 & 1792 & \\ \hline
\multicolumn{4}{|c|}{MaxPool2D 2x2}\\ \hline
    ConvBl2D & 64, 64, 256 & 36,928 & 37,440\\ \hline
    IdentBl2D x2 & 64, 64, 256 & 73,856 & 66,256\\ \hline
    ConvBl2D & 128, 128, 512 & 147,584 & 230,528 \\ \hline
    IdentBl2D x3 & 128, 128, 512 & 442,752 & 395,136 \\ \hline
    ConvBl2D & 256, 256, 1024 & 590,080 & 919,808 \\ \hline
    IdentBl2D x5 & 256, 256, 1024 & 2,950,400 & 2,627,840 \\ \hline
    ConvBl2D & 512, 512, 2048 & 2,359,808 & 3,674,624\\ \hline
    IdentBl2D x2& 512, 512, 2048 & 4,719,616 & 4,199,424 \\ \hline
    \multicolumn{4}{|c|}{GlobalAveragePooling2D} \\ \hline
    Dense & 100 & \multicolumn{2}{|c|}{204,900} \\ \hline
    \end{tabular}
\end{table}

\begin{figure}[t]
\begin{center}
\includegraphics[width=0.8\linewidth]{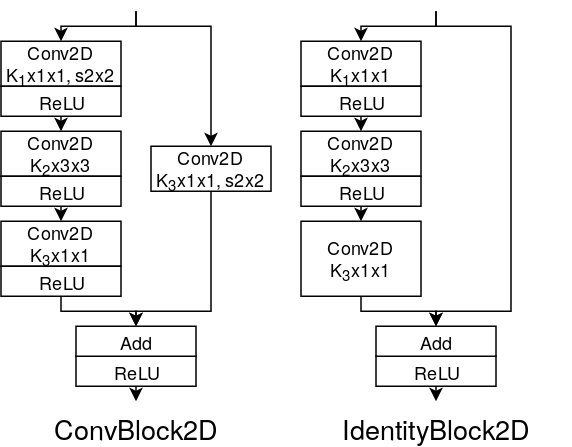}
\end{center}
    \caption{Structure of ConvBlock2D and IdentBlock2D. The term $K_1\times 3\times3$ denotes that a~convolutional layer includes $K_1$ filters of size $3\times3$. The term $2\times2$ denotes strides in both coordinates. The default strides are $1\times1$. A~batch normalization block \cite{IofSze15} is after each convolutional layer. }
    \label{fig:resblocks}
\end{figure}

\subsection{Problem 2. VGG-16 classifier} 
\label{sec:vgg-cifar100} 

\paragraph{Architecture} 
The original VGG architecture \cite{1409.1556} was designed to process large images. The architecture used here is based on VGG-CIFAR \cite{Liu2015VeryDC}. It was optimized to process smaller images. It has a~different number of dense layers, their sizes are different, and it contains batch normalization layers. Originally, VGG contains two dense layers with 4096 neurons and 1000 output neurons (the original VGG had been trained on ImageNet). VGG-CIFAR contains only one dense layer with 512 neurons and 100 output neurons (the number of classes in CIFAR-100). Additionally, as compared to the architecture from \cite{Liu2015VeryDC}, we resigned from drop-out after convolutional layers. That modification leads to a~noticeable improvement in performance on the test set. 


The activation function in each convolutional layer is ReLU. After each convolutional and dense layer there is a~batch normalization layer. A~detailed description of the architecture is presented in Tab.~\ref{tab:baseline_vgg}. The network has 15,028,644 trained parameters, including 14,710,464 in convolutional layers.


\begin{table}[]
\caption{VGG-16 architecture. 64x3x3 denotes 64 filters with slices of size 3x3. Conv2D x2 denotes 2 layers one after another.}
\label{tab:baseline_vgg}
\centering
\begin{tabular}{|c|c|c|}
\hline
\multicolumn{3}{|c|}{VGG-16}           \\ \hline \hline
Layer/Block & Size    & Parameters \\ \hline
Conv2D x2    & 64x3x3  & 38,720 \\ \hline
\multicolumn{3}{|c|}{MaxPool2D 2x2}       \\ \hline
Conv2D x2    & 128x3x3 & 221,440 \\ \hline
\multicolumn{3}{|c|}{MaxPool2D 2x2}       \\ \hline
Conv2D x3    & 256x3x3 & 1,475,328 \\ \hline
\multicolumn{3}{|c|}{MaxPool2D 2x2}       \\ \hline
Conv2D x3    & 512x3x3 & 5,899,776 \\ \hline
\multicolumn{3}{|c|}{MaxPool2D 2x2}       \\ \hline
Conv2D x3    & 512x3x3 & 7,079,424 \\ \hline
\multicolumn{3}{|c|}{MaxPool2D 2x2}       \\ \hline
Dense, ReLU  & 512 & 262,656 \\ \hline
\multicolumn{3}{|c|}{Dropout 0.5} \\ \hline
Dense, softmax& 100 & 51,300 \\ \hline
\end{tabular}
\end{table}

\paragraph{Training} 
The network was trained with the use of the NAG algorithm. The momentum decay factor was set to 0.9. The step-size was evolving according to the formula 
\begin{equation}
    \beta_t = \beta_0(0.5^{\floor{t/20}}), 
\end{equation}
where $t$ is the epoch index, $\beta_0=0.1$ is the initial step-size. The whole training lasted for 200 epochs. The loss function was categorical cross-entropy with L2 regularization of weights (with the coefficient of 0.0005). Minibatches of size 128 were used for the training. 


\subsection{Problem 3. Autoencoder 1} 
\label{sec:ae1} 

\paragraph{Architecture} 
The first autoencoder is taken from \cite{2019wawrzynski}. It contains 6 layers: 2 convolutional, 2 dense, and 2 layers with transposed convolution. Its architecture is presented in detail in Tab.~\ref{tab:ae1}. The network has 4,204,435 trained parameters, including 5,472 in convolutional layers and those with transposed convolution. 


\begin{table}[]
    \caption{Autoencoder 1. 16x3x3 denotes 16 filters with slices of size 3x3. Conv2DT denotes transposed convolution.} 
    \label{tab:ae1}
\centering
    \begin{tabular}{|c|c|c|}
    \hline
    \multicolumn{3}{|c|}{Autoencoder 1}           \\ \hline \hline
    Layer & Size    & Parameters \\ \hline
    Conv2D, ReLU & 16x3x3  & 448 \\ \hline
    \multicolumn{3}{|c|}{MaxPool2D 2x2}       \\ \hline
    Conv2D, ReLU & 16x3x3 & 2,320 \\ \hline
    Dense  & 512 & 2,097,664 \\ \hline
    Dense, sigmoid  & 4096 & 2,101,248 \\ \hline
    Conv2DT, ReLU & 16x3x3 & 2,320 \\ \hline
    \multicolumn{3}{|c|}{UpScale2D 2x2}       \\ \hline
    Conv2DTranspose, sigmoid & 16x3x3 & 435  \\ \hline
    \end{tabular}
\end{table}

\paragraph{Training} 
The network was trained with the ADAM algorithm using the momentum decay factor of 0.9 and the step-size of 0.001. The minibatches size was 64. The loss function was the binary cross-entropy. 


\subsection{Problem 4. Autoencoder 2} 
\label{sec:ae2} 

\paragraph{Architecture} 
The second autoencoder is based on https://github.com/Puayny/Autoencoder-image-similarity. However, the convolutions in its decoder were replaced with a~transposed convolution. The architecture is presented in detail in~Tab.~\ref{tab:ae2}. 
The network contains 390,955 trained parameters, including 390,240 in (transposed) convolution filters. 

\begin{table}[]
    \caption{Autoencoder 2. 16x3x3 denotes 16 filters with slices of size 3x3. Conv2DT denotes transposed convolution.} 
    \label{tab:ae2}
    \centering
    \begin{tabular}{|c|c|c|}
    \hline
    \multicolumn{3}{|c|}{Autoencoder 2}           \\ \hline \hline
    Layer & Size    & Parameters \\ \hline
    Conv2D, ReLU & 16x3x3 & 448 \\ \hline
    Conv2D, ReLU & 32x3x3 & 4,640 \\ \hline
    Conv2D, ReLU & 64x3x3 & 18,496 \\ \hline
    \multicolumn{3}{|c|}{MaxPool2D 2x2}       \\ \hline
    Conv2D, ReLU & 128x3x3 & 73,856 \\ \hline
    Conv2D, ReLU & 64x3x3 & 73,792 \\ \hline
    \multicolumn{3}{|c|}{MaxPool2D 2x2}       \\ \hline
    Conv2D, ReLU & 32x3x3 & 18,464 \\ \hline
    Conv2D, ReLU & 16x3x3 & 4,624 \\ \hline
    Conv2D, sigmoid & 8x3x3 & 1,160 \\ \hline
    Conv2DT, ReLU & 16x3x3 & 1,168 \\ \hline
    Conv2DT, ReLU & 32x3x3 & 4,640 \\ \hline
    \multicolumn{3}{|c|}{UpSampling2D 2x2}       \\ \hline
    Conv2DT, ReLU & 64x3x3 & 18,496 \\ \hline
    Conv2DT, ReLU & 128x3x3 & 73,856 \\ \hline
    \multicolumn{3}{|c|}{UpSampling2D 2x2}       \\ \hline
    Conv2DT, ReLU & 64x3x3 & 73,792 \\ \hline
    Conv2DT, ReLU & 32x3x3 & 18,464 \\ \hline
    Conv2DT, ReLU & 16x3x3 & 4,624\\ \hline
    Conv2DT, sigmoid & 3x3x3 & 435 \\ \hline 
    \end{tabular}
\end{table}

\paragraph{Training} 
The network was trained with the ADAM algorithm using a~momentum decay factor of 0.9 and the step-size of 0.0005 (this was the largest value that assured stable learning of the network). The minibatches size was 64. The loss function was the binary cross-entropy. 


\subsection{Implementation and computational efficiency} 

Our experimental software has been written in Python/Tensorflow 2.0. DCT-Conv layers were implemented such that an additional DCT block was applied to produce filters of ordinary convolutional (or transposed convolution) layers. Switching off was implemented such that the weights were multiplied elementwise by mask tensors with appropriate number of entries set to zero. 

The real time overhead resulting from adding the DCT block to the convolutional layer is comparable to the computing time that the original layer takes. Therefore, our experiments with DCT-Conv networks lasted up to twice longer than with the original convolutional networks. This time could be significantly reduced if DCT-Conv layer was implemented as a~single block. 

\subsection{Results}

\def\ns{\!\!&\!\!}

\begin{table} 
\caption{Results. Each reported result is an~average registered on the test set after training. The average is over 5 runs. Left-most columns contain results for networks with CNN layers. Other columns contains results for networks with DCT-Conv layers with different probability of switching off weights. } 
\label{tab:results} 
\begin{center} 
Problem 1, ResNet50, all filters. Result reported: Accuracy 
\begin{tabular}{|l|l|l|l|l|l|l|} 
\hline 
CNN & $p=0$ \ns $p=0.3$ \ns $p=0.5$ \ns $p=0.7$ \ns $p=0.9$ \ns $p=0.97$ \\ 
\hline
0.6698 & 0.7127 & 0.7198 & 0.6845 & 0.7044 & 0.7067 & 0.6529 \\ 
\hline
\end{tabular} 
\\*[1ex]
Problem 1, ResNet50, only 3x3 filters. Result reported: Accuracy 
\begin{tabular}{|l|l|l|l|l|l|} 
\hline 
CNN & $p=0$ & $p=0.99$ & $p=0.999$ & $p=0.9999$ & $p=1$ \\ 
\hline
0.6698 & 0.6826 & 0.7331 & 0.7141 & 0.6415 & 0.3416 \\
\hline
\end{tabular} 
\\*[1ex]
Problem 2, VGG-16. Result reported: Accuracy
\begin{tabular}{|l|l|l|l|l|l|} 
\hline 
CNN & $p=0$ & $p=0.3$ & $p=0.5$ & $p=0.7$ & $p=0.9$ \\ 
\hline
0.7177 & 0.7221	& 0.7088 &0.7099	&0.6930	&0.6517 \\
\hline
\end{tabular} 
\\*[1ex]
Problem 3, Autoencoder 1. Result reported: MSE for pixels in $[0,1]$ 
\begin{tabular}{|l|l|l|l|l|l|} 
\hline 
CNN & $p=0$ & $p=0.3$ & $p=0.5$ & $p=0.7$ & $p=0.9$ \\ 
\hline 
0.00136 & 0.00126 & 0.00132 & 0.00138 & 0.00151 & 0.00283 \\ 
\hline
\end{tabular} 
\\*[1ex]
Problem 4, Autoencoder 2. Result reported: MSE for pixels in $[0,1]$
\begin{tabular}{|l|l|l|l|l|l|} 
\hline 
CNN & $p=0$ & $p=0.3$ & $p=0.5$ & $p=0.7$ & $p=0.9$ \\ 
\hline
0.00153	& 0.00118	& 0.00135	& 0.00149 &	0.00188 & 0.00381 \\  
\hline
\end{tabular} 
\end{center} 
\end{table}

All the experiments were performed according to the following pattern: 
\begin{enumerate} 
\item The original neural network is trained. 
\item The convolutional layers of the network are replaced with DCT-Conv layers. Its trained DCT coefficients are initialized as the components of the filters were originally initialized. The network is retrained from the beginning according to the original regime. 
\item The DCT coefficients of the DCT-Conv layers are reinitialized, and  some of them are switched off, i.e. set equal to zero and excluded from the training. They are switched off on random, each with probability $p$. Technically, they are assigned random real numbers, sorted according to those numbers, and first $p100\%$ of them are switched off. Once again, the network is retrained from the beginning according to the original regime. 
\end{enumerate} 
Note that the network architecture, as well as its training regime, has been optimized with respect to its performance when it is used without any further modifications. We do not change any of these when replacing the original convolutional layers with DCT-Convs. 

The results are presented in Tab.~\ref{tab:results}. Each second row in Tab.~\ref{tab:results} presents performance on the test set. Each number averages 5 runs. For problems 1 and 2, the performance is expressed with accuracy. For problems 3 and 4, the performance is expressed with a~mean square error. Every left column shows the performance of the original convolutional neural network. Other columns demonstrate the performance of the network with its convolutional layers replaced by DCT-Convs and their DCT coefficients switched off with a~different probability, $p$. 

\paragraph{Problem 1} 
Here we can observe that the network with our proposed layers outperforms the original one if only $p \leq 0.9$. 

A~debatable issue for this problem is how to treat $1\times1$ convolutions. At first, we joined them over 16 consecutive channels and performed IDCT on such packs. In another experiment we leave the original convolutional layers with $1\times1$ filters, and the convolutional layers with $W\times H$ filters for $W,H>1$ are replaced with DCT-Convs. The result is stunning: Even if DCT coefficients of such layers are switched off with the probability $p=0.999$, the network performs better than the original one. Let us consider a~filter with $3\times3$ slices and $128$ input channels. It has $3\cdot3\cdot128=1152$ DCT coefficients. When they are switched off with the probability $p=0.999$, some filters are switched off entirely, and each of the others has only a~few slices that are in fact operational. This happens to be enough for a~fairly good performance of the network. However, it is not true that $3\times3$ filters are useless in this network. When they are all switched off ($p=1$), the network performs poorly. 

\paragraph{Problem 2} 
Here we can observe that the network with our proposed layers outperforms the original one if only $p$ is very close to zero. However, the accuracy on the test set decreases rather slowly with growing $p$. 

\paragraph{Problem 3} 
Here we can observe that the network with our proposed layers outperforms the original one if only $p < 0.5$. 

\paragraph{Problem 4} 
Here we can observe that the network with our proposed layers outperforms the original one if only $p \leq 0.5$. 

\subsection{Summary} 

In all problems analyzed in the above empirical study, a~network with our proposed DCT-Conv layers performed better than the original network with convolutional layers. Switching DCT coefficients off in DCT-Conv layers led to the deterioration of performance. However, the performance decreased rather slowly with the growing probability of switching off. If less than about 50\% DCT coefficients were switched off, the network still performed better than the original one. However, in the case of the ResNet50 classifier (Problem 1) 90\% DCT coefficients could be switched off without deterioration of performance. If only convolutional layers with $3\times3$ filters in ResNet50 were replaced with DCT-Convs, then 99.9\% of DCT coefficients could be switched off without deterioration of performance. 

\section{Conclusions and future work}
\label{sec:conclusions} 

In this paper DCT-Conv layer was introduced --- a~layer in which trained weights are DCT coefficients of spatial filters. The DCT-Conv layer realizes the idea that a~sufficiently rich set of spacial filters can have sparse frequency representation. 

In four experiments with benchmark convolutional neural networks, it was demonstrated that the networks with their convolutional layers replaced by DCT-Conv layers outperform the original networks even if large part, about 50\%, of their DCT coefficients are switched off (set equal to zero and excluded from training). In some cases that part could be significantly larger. 

In this paper we considered switching off the DCT coefficients of the DCT-Conv layer on random with equal probability for each layer. We plan to investigate strategies of determining those probabilities for different layers and their specific DCT coefficients. Also, optimization of the DCT-Conv layers shape is a~curious topic of future research. 

\section*{Acknowledgment}

We gratefully acknowledge the support of NVIDIA Corporation with the donation of the Titan X Pascal GPU used for this research.

\bibliographystyle{IEEEtranN}


\end{document}